\documentclass[12pt, twocolumn]{article}
\usepackage[left=0.6in, right=0.6in, top=0.8in, bottom=0.8in]{geometry}
\usepackage{graphicx}
\usepackage{amsmath, amssymb}
\usepackage{hyperref}
\usepackage[numbers,sort&compress]{natbib}
\usepackage{titlesec} 
\usepackage{ragged2e}
\usepackage[T1]{fontenc}
\usepackage{dblfloatfix}
\usepackage{placeins}
\usepackage{booktabs, tabularx, makecell}
\usepackage{xurl}
\usepackage[font=small,labelfont=bf]{caption}

\titleformat{\section}{\large\bfseries}{\thesection}{1em}{}
\titleformat{\subsection}{\normalsize\bfseries}{\thesubsection}{1em}{}
\newcounter{numquote}

\title{Training-Free Active Learning Framework in Materials Science with Large Language Models}
\author{
    Hongchen Wang$^{1,\dagger}$, Rafael Espinosa Casta\~neda$^{1,5,\dagger}$,
    Jay R. Werber$^{2}$, Yao Fehlis$^{3}$,\\
    Edward Kim$^{1,4,*}$, Jason Hattrick-Simpers$^{1,5,6,*}$
}
\date{}
\begin{document}

\justifying
\twocolumn[
\maketitle
\begin{center}
    {\small 
    $^{1}$Department of Materials Science and Engineering, University of Toronto, ON, Canada\\
    $^{2}$Department of Chemical Engineering, University of Toronto, ON, Canada\\
    $^{3}$KUNGFU.AI, Austin, Texas, United States\\
    $^{4}$Cohere, Toronto, ON, Canada\\
    $^{5}$Vector Institute, Toronto, ON, Canada\\
    $^{6}$Acceleration Consortium, Toronto, ON, Canada\\
    $^{\dagger}$These authors contributed equally to this work.\\
    $^{*}$Corresponding authors: edwardsoo.kim@mail.utoronto.ca, jason.hattrick.simpers@utoronto.ca
    }
\end{center}
    
\begin{abstract}
Active learning (AL) accelerates scientific discovery by prioritizing the most informative experiments, but traditional machine learning (ML) models used in AL suffer from cold-start limitations and domain-specific feature engineering, restricting their generalizability. Large language models (LLMs) offer a new paradigm by leveraging their pretrained knowledge and universal token-based representations to propose experiments directly from text-based descriptions. Here, we introduce an LLM-based active learning framework (LLM-AL) that operates in an iterative few-shot setting and benchmark it against conventional ML models across four diverse materials science datasets. We explored two prompting strategies: one using concise numerical inputs suited for datasets with more compositional and structured features, and another using expanded descriptive text suited for datasets with more experimental and procedural features to provide additional context. Across all datasets, LLM-AL could reduce the number of experiments needed to reach top-performing candidates by over 70\% and consistently outperformed traditional ML models. We found that LLM-AL performs broader and more exploratory searches while still reaching the optima with fewer iterations. We further examined the stability boundaries of LLM-AL given the inherent non-determinism of LLMs and found its performance to be broadly consistent across runs, within the variability range typically observed for traditional ML approaches. These results demonstrate that LLM-AL can serve as a generalizable alternative to conventional AL pipelines for more efficient and interpretable experiment selection and potential LLM-driven autonomous discovery.
\end{abstract}
\vspace{1em}
]
\clearpage

\section{Introduction}

Material discovery and process optimization often requires exploring large design spaces, where exhaustive trial-and-error is impractical due to time and cost constraints \cite{Axelrod2022, Merchant2023, Sadeghi2025}. Active learning (AL) has therefore become a critical tool in experimental science to improve data efficiency and guide experimental decision making \cite{Lookman2019, Kusne2020, Wang2022}. By prioritizing the most informative experiments, AL accelerates discovery while reducing the total number of trials required. With advances in laboratory automation, AL has become the core component of self-driving laboratories (SDLs), enabling closed-loop selection of experiments to efficiently guide exploration and discovery \cite{Tobias2025, Hickman2025, Li2025}. Implementations using Bayesian optimization and related AL frameworks have repeatedly demonstrated order-of-magnitude improvements in experimental efficiency, often identifying optimal material candidates with up to 10 to 20 times fewer experimental iterations compared to unguided or random screening approaches \cite{Kusne2020, adesiji2025benchmarkingselfdrivinglabs, Rohr2020-mu, Suvarna2024}. 

Although traditional machine learning (ML)-based AL has shown significant progress, its scalability and versatility remain limited. Classical surrogate models rely heavily on task-specific feature engineering and sufficient labeled data to generate reliable predictions. In early rounds of experimentation, when only a small number of data points are available, these models often exhibit low predictive accuracy and unreliable uncertainty estimates, leading to the "cold start" problem and inefficient exploration of the design space \cite{yuan-etal-2020-cold, bayer2025activellmlargelanguagemodelbased, bar2024activelearningclassifierimpact}. In addition, hyperparameter tuning for these surrogate models is highly brittle when data is scarce, often yielding extreme or suboptimal parameter settings \cite{Han2021, Pardakhti_2021, Manzhos_2022}. A deeper limitation of traditional AL models lies in their lack of generalizability across domains \cite{lowell-etal-2019-practical, xie2025darwin15largelanguage}. Most classical models operate in problem-specific feature spaces, requiring careful feature engineering or model tuning for each new task \cite{lowell-etal-2019-practical, xie2025darwin15largelanguage}. For example, a Gaussian process regression (GPR) model trained on polymer formulation data cannot be applied to an alloy design problem without reworking the input representations and hyperparameters, which limits its transferability and slows its iteration across diverse experimental domains.

Human experts leverage decades of accumulated knowledge from the literature and prior experience to make informed experimental decisions, even with sparse data \cite{Charness2008}. Large language models (LLMs) can extend this knowledge-driven reasoning into computational frameworks \cite{prabhakar2025omnisciencedomainspecializedllmscientific, Zhang2025}. Drawing on vast amounts of scientific corpora and domain knowledge, LLMs can identify patterns and relationships between material descriptors and properties that can be effectively used to guide exploration \cite{xia-etal-2025-selection, Jiang2025}. Studies have shown that LLMs can mitigate the cold start problem, providing meaningful experimental guidance even in the earliest iterations when labeled data are scarce \cite{bayer2025activellmlargelanguagemodelbased, senthilkumar2024largelanguagemodelsimprove, Ciss__2025}. Furthermore, by operating in a universal token-based feature space, LLMs can generalize across diverse scientific domains without extensive feature engineering, enabling more adaptive and transferable AL workflows \cite{Zhang2025, Pathan2025, boiko2023emergentautonomousscientificresearch}. 

Previous studies have demonstrated their property prediction capabilities in different scientific domains using few-shot in-context learning (ICL) \cite{WANG20251612, LIU202523, D4SC03921A}. In particular, pretrained LLMs can be adapted through carefully designed prompts to perform property prediction in zero-shot or few-shot settings, leveraging their broad prior knowledge for domain-specific inference without any parameter updates, making them well-suited to guide experiment selection and support AL workflows \cite{WANG20251612, LIU202523, D4SC03921A}. However, employing LLMs for scientific AL also presents unique challenges. Their non-deterministic nature, even under controlled decoding conditions, introduces variability that requires stability analysis to ensure reproducibility \cite{he2025nondeterminism}. They can also be sensitive to prompt phrasing and occasionally generate inconsistent or non-physical outputs when extrapolating beyond the scope of their training data \cite{WANG20251612}. Overall, the performance and limitations of LLMs as surrogate models for iterative design and optimization in materials science remain underexplored, particularly in comparison to traditional ML approaches.

In this work, we designed a pool-based AL workflow to evaluate the capabilities of LLMs as surrogate models for guiding experimental design. In this approach, the model iteratively queries a fixed pool of unlabeled data to identify the next most informative experiment. We benchmarked an LLM-driven AL framework (LLM-AL) against four widely used traditional ML approaches: random forest regressor (RFR), eXtreme gradient boosting (XGBoost), Gaussian process regressor (GPR), and Bayesian neural network (BNN). The evaluation was conducted across four heterogeneous datasets spanning key materials science domains, including alloy design, polymer nanocomposites, perovskite, and membrane optimization. Our results show that, even without fine-tuning, the LLM-AL workflow achieves competitive or superior performance, consistently reaching optimal candidates with fewer iterations than traditional ML models. By comparing concise, structured prompts with more descriptive, context-rich prompts, we show how prompt design can be tailored to different types of experimental data. We also examined the stability boundaries of LLM-AL given the inherent non-determinism of LLMs, finding that its variability across repeated runs and random seeds remained comparable to that of traditional ML models. The results suggest that LLMs can serve as flexible, tuning-free, and generalizable tools in AL pipelines, providing effective experimental guidance across diverse materials science domains.

\section{Results}

\subsection{Performance and Reproducibility of LLM-based Active Learning}

\begin{figure*}[!ht]
\centering
\includegraphics[width=\linewidth]{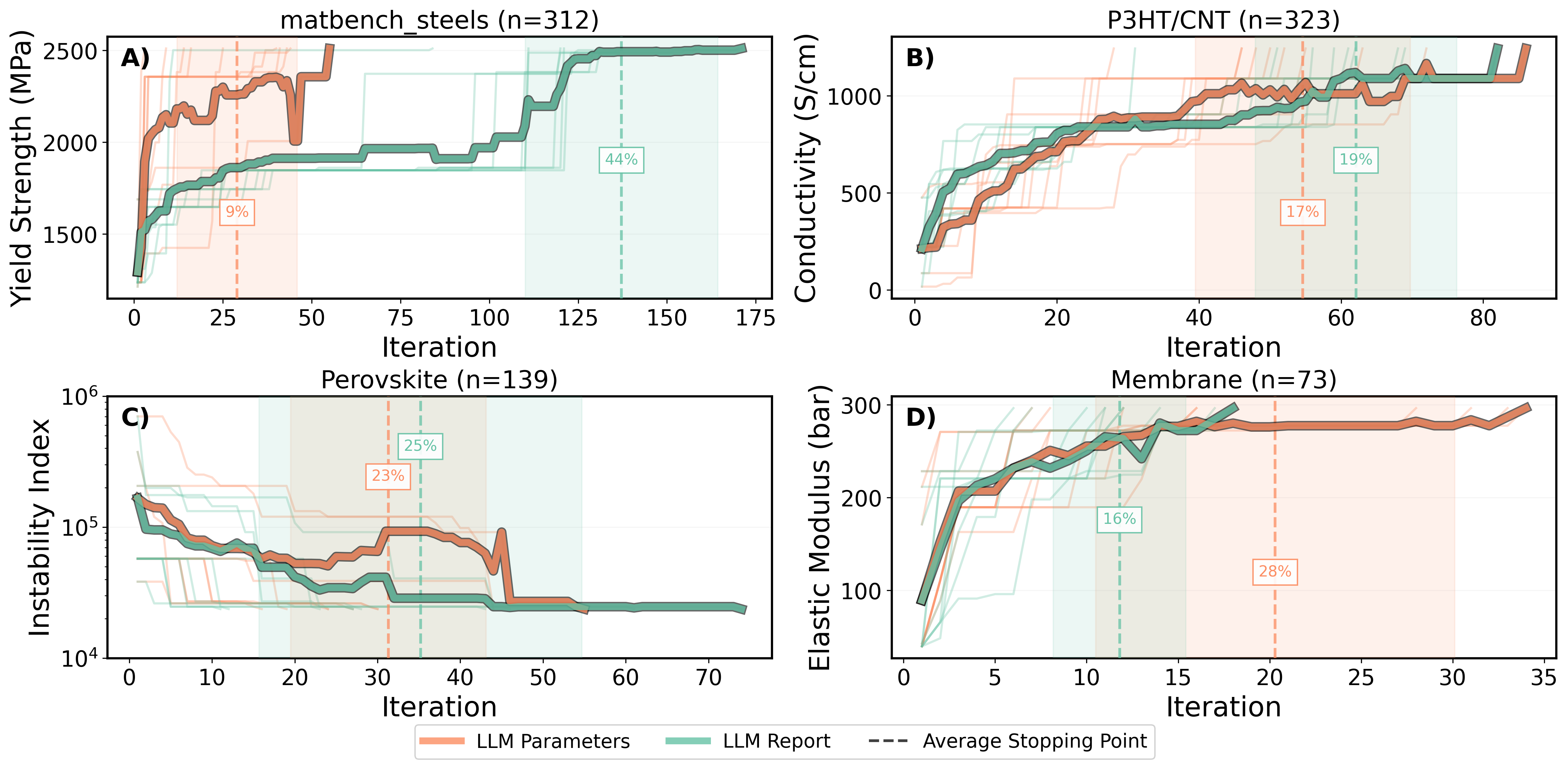}
\caption{The trajectories of the running best performance for the LLM-AL approach across four datasets: (A) matbench\_steels, (B) P3HT/CNT, (C) Perovskite, and (D) Membrane. Each dataset has a maximization goal for its target property, except the Perovskite dataset which has a minimization goal. Orange lines indicate runs using the parameter-format input prompt, and green lines indicate runs using the report-format prompt. For each setting, five random seeds and five repeated runs were performed, for a total of 10 runs. The thicker lines represent the average running best trajectory across the 10 runs. Vertical dashed lines mark the mean iteration at which the stopping criterion is first reached, with shaded regions showing one standard deviation. The annotated percentages indicate the fraction of the dataset used to reach the stopping point.}
\label{fig:AL_llm_trajectory.png}
\end{figure*}

Figure~\ref{fig:AL_llm_trajectory.png} shows how the LLM-AL framework performs across four representative datasets (i.e., matbench\_steels, P3HT/CNT, Perovskite, and Membrane). These trajectories demonstrate the ability of the LLM to iteratively propose experimental suggestions given prior observations and identify optimal targets with minimal data usage. Moreover, it is important to understand how the structure of textual prompts influences model performance for using LLMs in materials discovery and optimization workflows. Unlike traditional ML models that rely on numerical features, LLMs interpret text-based representations, meaning that the phrasing, context, and level of detail in each prompt can significantly impact their responses. Thus, we compared two prompting strategies, a concise parameter-format, where inputs are structured as feature–value pairs, and a narrative report-format, where inputs are rewritten into experimental descriptions, to evaluate how the structure and complexity of textual representations influence the model’s efficiency and convergence behavior.

\begin{figure*}[!ht]
\centering
\includegraphics[width=\linewidth]{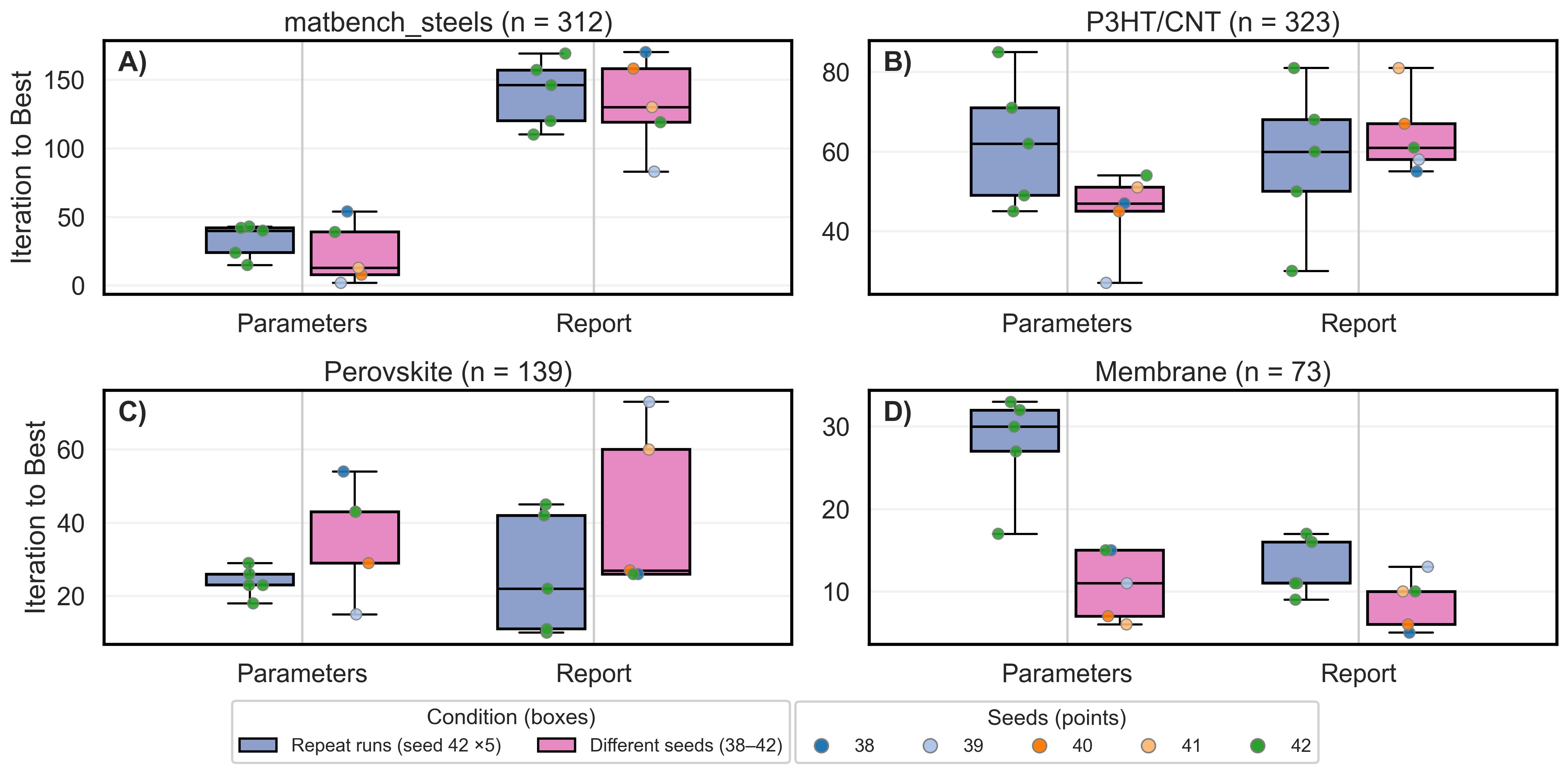}
\caption{The number of iterations required for the LLM-AL approach to reach the optimal target across four datasets: (A) matbench\_steels, (B) P3HT/CNT, (C) Perovskite, and (D) Membrane. Blue boxplots represent repeated runs using the same random seed (seed 42, five repeats), while pink boxplots represent runs with different random seeds (seeds 38–-42). Individual points show the results for each run, color-coded by seed. Each dataset is divided into two groups corresponding to the parameter-format input prompts and report-format input prompts.}
\label{fig:AL_llm_variation.png}
\end{figure*}

We observed that LLM-AL consistently converges to the optimal target using less than 30\% of the data, except for the matbench\_steels dataset with report-format prompts, which required 44\% of the data on average. This dataset contains the largest number of input features, represented as compositions. Converting these high-dimensional inputs into a descriptive report format may not add meaningful information and could even dilute key details, leading to increased computational costs and slower convergence \cite{du2025contextlengthhurtsllm, kusano2024longerpromptsbetterprompt}. Long prompts can make it harder for the model to focus on the most important features, especially when key details appear in the middle of the text, known as the "lost-in-the-middle" effect \cite{liu2023lostmiddlelanguagemodels, modarressi2025nolimalongcontextevaluationliteral}. In contrast, the other datasets involve fewer and primarily experimental parameters. The Membrane dataset shows the most notable improvement with the report-format prompt, likely because it contains procedural descriptors that become more informative when expanded into narrative form, enabling the LLM to extract hidden relationships and make better suggestions \cite{kusano2024longerpromptsbetterprompt}. As this dataset is entirely original and unpublished, it is highly unlikely to have appeared in any LLM training corpus, lending further confidence that the observed improvement arises from contextual reasoning rather than memorization.

These results suggest that LLM-AL can serve as a robust and generalizable tool for guiding experimental design across diverse domains, spanning metals and polymers, compositional and experimental tasks, and both maximization and minimization objectives. However, the prompting strategy must be tailored to the dataset. For datasets with concise, procedural inputs, detailed report-format prompts can help capture underlying information and enhance reasoning. In contrast, datasets containing many independent variables (e.g., compositions) favor parameter-format prompts, which can mitigate unnecessary computation and performance degradation.

As LLMs are non-deterministic models, their responses can vary even when using identical input prompts and a temperature of 0, meaning that the same prompt can produce different outputs on different runs \cite{he2025nondeterminism}. This inherent randomness introduces an additional layer of uncertainty, making it important to carefully evaluate the reproducibility of the LLM-AL framework before deployment in experimental workflows. By distinguishing between variability due to the model’s inherent stochasticity and that due to initial data selections, we can better understand the stability boundaries of LLM-guided active learning. Figure~\ref{fig:AL_llm_variation.png} compares the number of iterations required by LLM-AL to reach the optimal target under two conditions: (1) repeated runs with a fixed random seed and (2) runs initialized with different random seeds. This design enables the separation of stochasticity arising purely from the LLM’s internal randomness versus the variability introduced by changes in initial conditions.

\begin{figure*}[!ht]
\centering
\includegraphics[width=\linewidth]{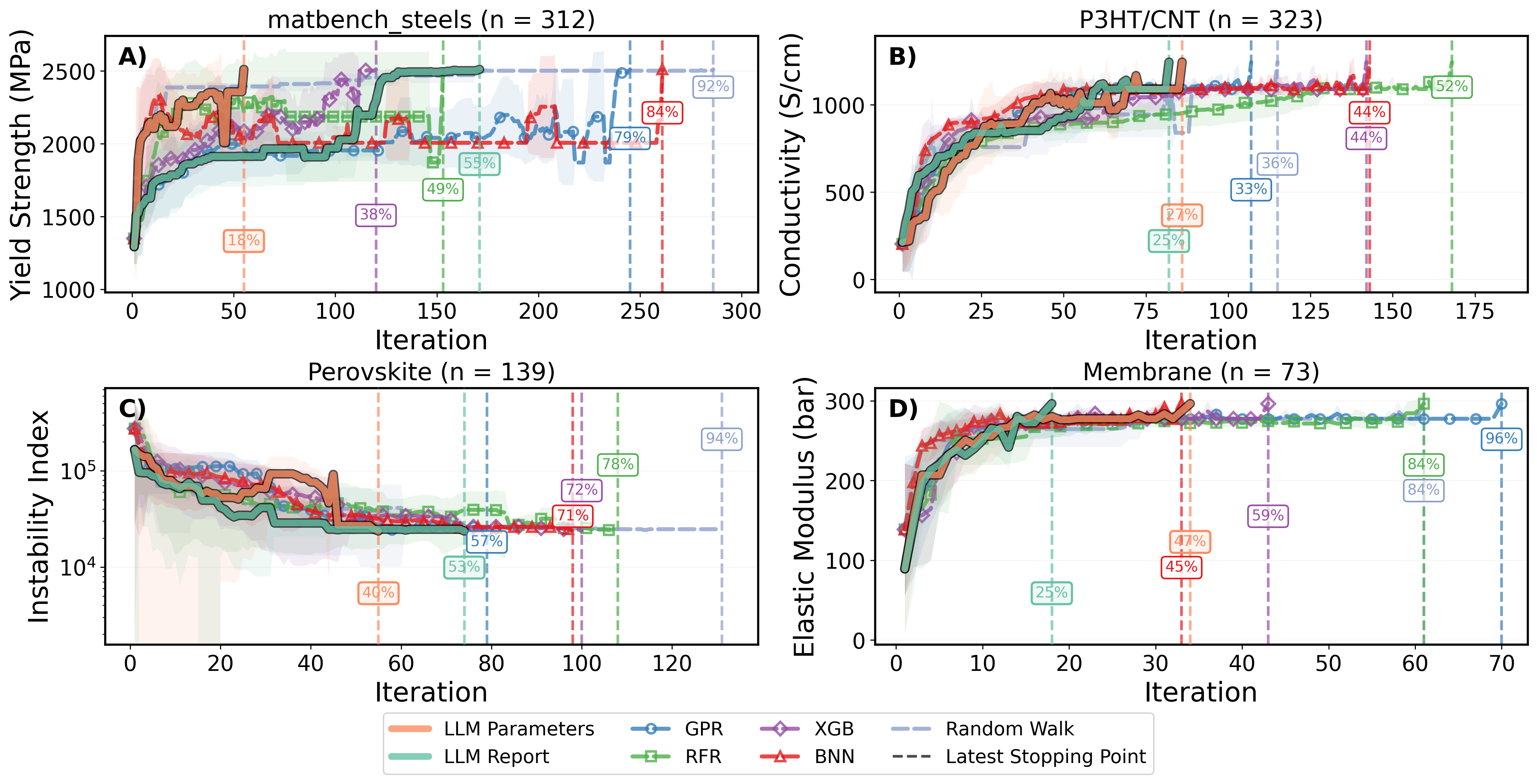}
\caption{The trajectories of the running best performance for LLM-AL and traditional ML models across four datasets: (A) matbench\_steels, (B) P3HT/CNT, (C) Perovskite, and (D) Membrane. Bold orange and green lines represent LLM-AL runs using the parameter-format and report-format prompts, respectively. Other colored dashed lines represent traditional ML models and random walk: GPR (blue), RFR (green), XGB (purple), BNN (red), and random walk (pale blue). For LLM-AL, shaded regions indicate the variability across 10 runs, consisting of five repeated runs using a fixed random seed (42) and five distinct random seeds (38--42) used to define the initial data pool. 
For traditional ML models, shaded regions capture the combined variability across five random seeds (38--42) and a range of UCB trade-off values ($\alpha = 0$ to $5$), which control the exploration--exploitation balance during experiment selection. Vertical dashed lines indicate the latest iteration where the stopping criteria were met, and annotated percentages indicate the fraction of the total dataset required to reach that point.}
\label{fig:AL_trajectory.png}
\end{figure*}

We observed that the variance of repeated runs is of magnitude similar to the variance across different seeds, suggesting that most of the variability is intrinsic to the LLM itself rather than arising from the data initialization. The report-format prompts generally exhibited greater variance than the parameter-format prompts, likely because of their increased length and complexity. Longer, narrative-style prompts may amplify sensitivity to small changes in token generation, resulting in more divergent optimization trajectories \cite{guan2025ordereffectinvestigatingprompt}. Although this variability is greater than what is generally observed in traditional ML models, it remains within a practical and acceptable range, especially when multiple runs are averaged or replicated. Importantly, in all datasets, the final performance levels remain consistently high, indicating that while LLM-AL may take slightly different paths to convergence, the ultimate optimization outcomes are reliable. A more detailed benchmark of performance and variability against traditional ML models will be presented in Section~\ref{sec:Performance Benchmarking against Traditional ML Models}.

\subsection{Performance Benchmarking against Traditional ML Models}
\label{sec:Performance Benchmarking against Traditional ML Models}

Figure~\ref{fig:AL_trajectory.png} compares the performance of LLM-AL with traditional ML models across the four datasets to test whether LLM-AL can achieve similar or faster convergence than conventional AL strategies. Conventional AL approaches rely on predictive models trained on data seen so-far and explicit uncertainty estimation to guide data selection, while LLMs additionally rely on their pretrained knowledge and text-based reasoning to infer promising directions directly from prior observations. The performance of traditional ML baselines was evaluated across a range of exploration–exploitation balance parameters (i.e., $\alpha$) within the UCB acquisition function, to assess the models' sensitivity to this trade-off. This enables a fair comparison not only of convergence efficiency but also of the stochastic behavior of LLMs and the parameter-dependent variability of traditional AL models.

\begin{figure*}[!ht]
\centering
\includegraphics[width=\linewidth]{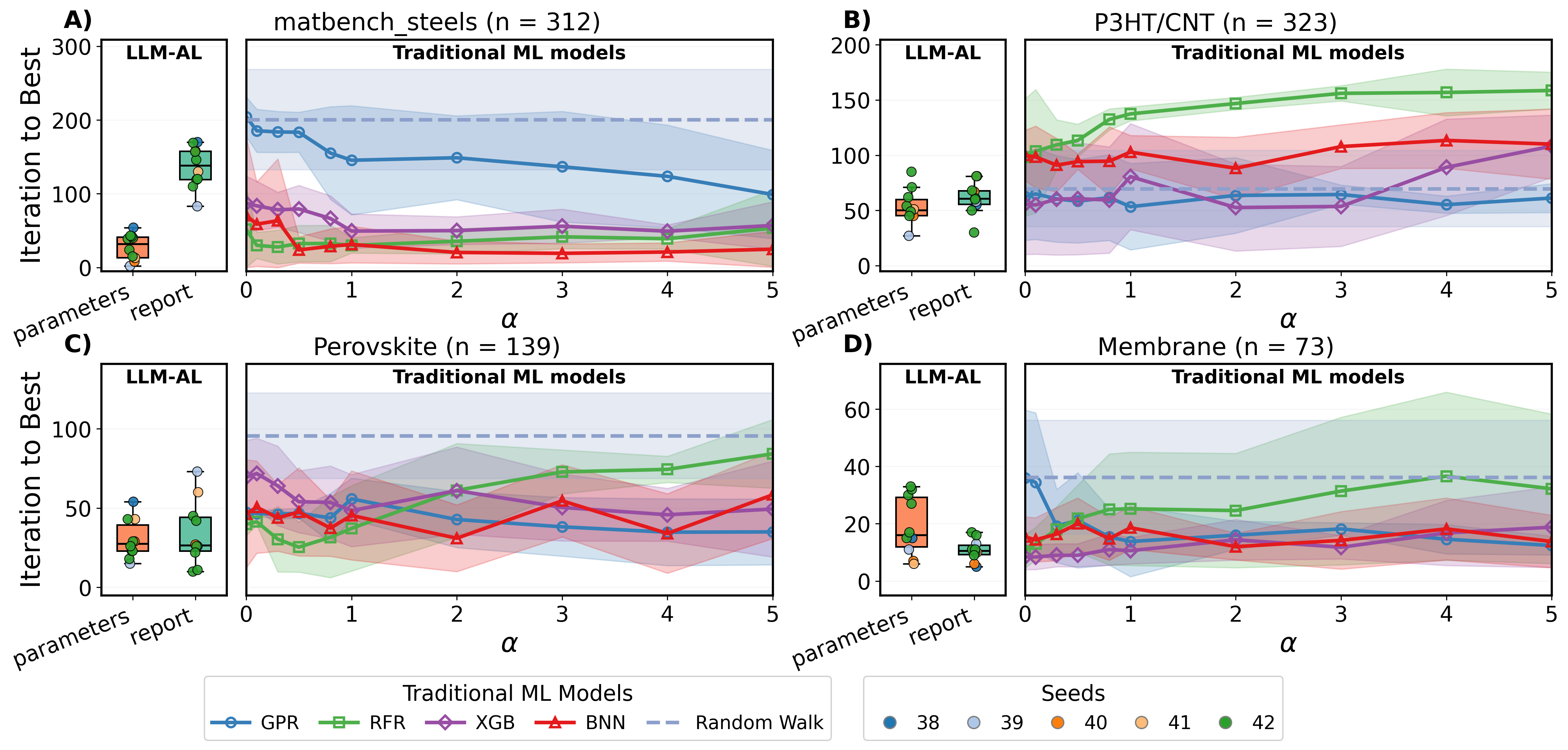}
\caption{Iterations required to reach the best-performing candidate for LLM-AL compared to traditional ML models across four datasets: A) matbench\_steels, B) P3HT/CNT, C) Perovskite, and D) Membrane. For LLM-AL (left panels), the box plots show the distribution of iterations required to reach the maximum target value across five random seeds (38–42) and five repeated runs at seed 42. For traditional ML models (right panels), the line plots show the mean iterations required to reach the maximum target value as a function of the UCB exploration--exploitation trade-off parameter, $\alpha$, with shaded regions representing the standard deviation across the same five seeds.}
\label{fig:AL_performance.png}
\end{figure*}

For the matbench\_steels dataset, traditional ML models such as GPR, RFR, XGB, and BNN required between 38–-81\% of the dataset to reach the best-performing candidate in their worst-case scenarios. In contrast, LLM-AL converged much earlier with the parameter-format prompt, identifying the best candidate using only 18\% of the data. However, when using the report-format prompt, the performance dropped noticeably, with the LLM requiring substantially more data (55\%) and underperformed some traditional ML models. This decline is likely due to the high-dimensional compositional features of the dataset, where descriptive text adds unnecessary complexity. The resulting long prompts increase computational costs and can distract the LLM’s attention, causing slower convergence. Traditional ML models, on the other hand, may be better suited to handle large parameter spaces and complex features.

Across the remaining datasets, P3HT/CNT, Perovskite, and Membrane, LLM-AL consistently demonstrates superior performance over traditional ML-based AL strategies. Both parameter-format and report-format prompts for LLM-AL converge to the optimal candidate using fewer iterations compared to traditional ML models, even in their worst-case scenarios. This highlights the ability of LLMs to leverage their inherent prior knowledge and generalization capabilities to navigate experimental search spaces without explicit model retraining or hyperparameter tuning. 

For the Membrane dataset, which features highly procedural experimental parameters, the report-format prompt is particularly effective, converging significantly faster than all traditional ML models. This suggests that LLMs can extract implicit knowledge from rich procedural descriptions, leading to more efficient exploration and optimization. Overall, these results indicate that LLM-AL is not only competitive but often superior to conventional uncertainty-based active learning strategies, particularly when the search space is smaller or when descriptive experimental context is available to enhance decision-making.

To further explore the variability in performance, Figure ~\ref{fig:AL_performance.png} compares the number of iterations required for LLM-AL and traditional ML-based AL approaches to reach the optimal target across the four datasets. The left panels show the LLM-AL's performance as distributions across five random seeds, while the right panels show traditional ML performance as a function of the UCB exploration–exploitation parameter, $\alpha$. Smaller $\alpha$ values favor exploitation and lead to slower exploration, while larger $\alpha$ values encourage exploration but may cause inefficient sampling. Even at their optimal $\alpha$, traditional ML models generally require as much or more data to reach the same performance level as LLM-AL. The distributions for LLM-AL are relatively compact, indicating consistent performance across random seeds. In contrast, traditional ML models exhibit more variable convergence behavior that strongly depends on the choice of $\alpha$.

\begin{figure*}[!ht]
\centering
\includegraphics[width=\linewidth]{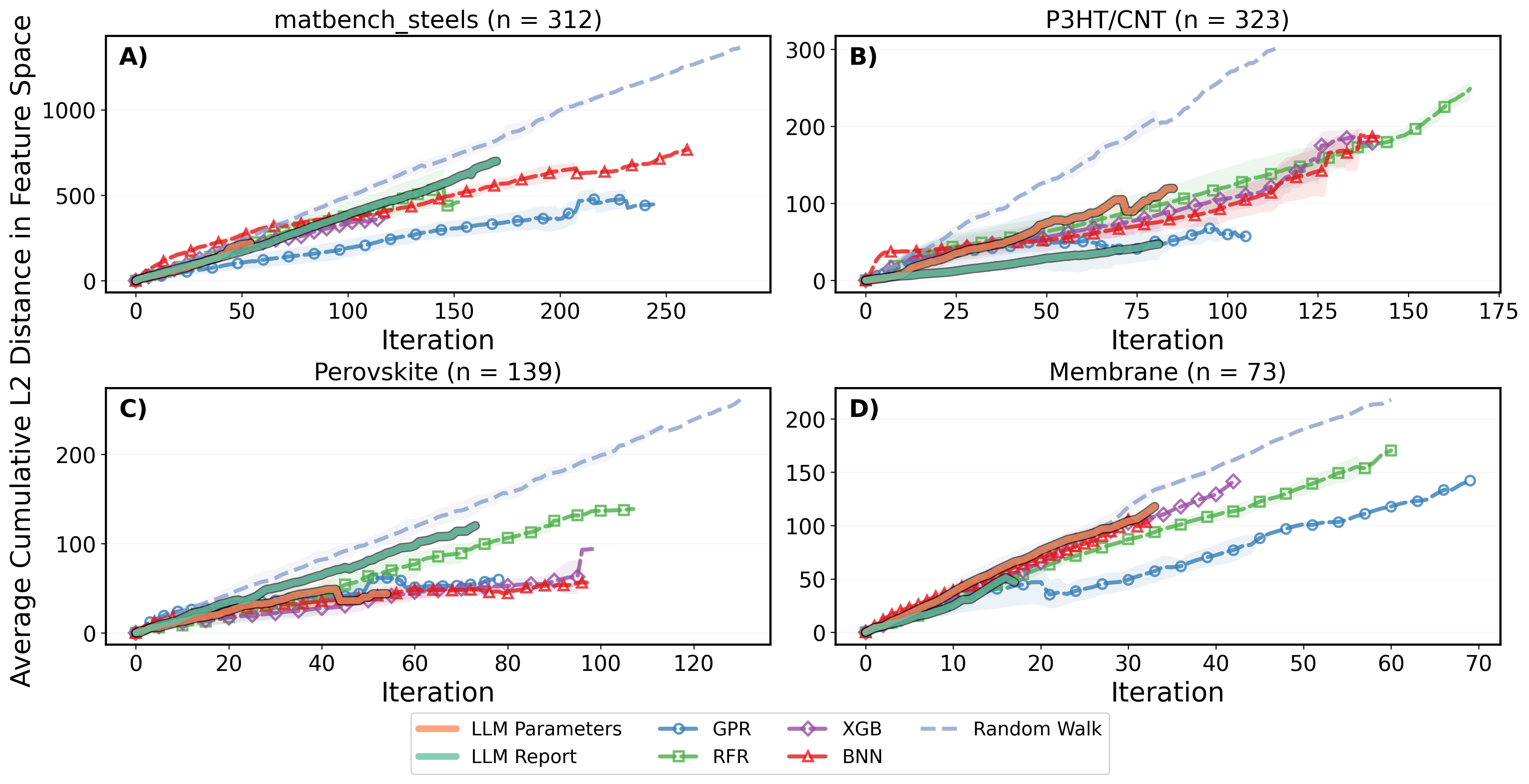}
\caption{Average cumulative L2 distance traveled in standardized feature space as a function of iteration across four datasets: A) matbench\_steels, B) P3HT/CNT, C) Perovskite, and D) Membrane. Each line represents the mean cumulative L2 trajectory length across multiple random seeds, with shaded regions showing the standard deviation. The figure compares LLM-AL using parameter and report prompting formats against traditional ML models (GPR, RFR, XGB, BNN) and a Random Walk baseline.}
\label{fig:AL_distance.png}
\end{figure*}

A key insight is that LLM-AL can achieve strong and consistent performance with minimal tuning and feature engineering, unlike traditional ML models, where performance is highly sensitive to hyperparameters (e.g., $\alpha$ in the UCB acquisition function). While prompt design represents a form of hyperparameter selection, its conceptual simplicity contrasts with the multi-dimensional, task-specific optimization required by traditional ML pipelines, and it can be more intuitively adapted based on the nature of the dataset (e.g., compositional vs. procedural). This highlights the robustness of LLM-AL in scenarios where the exploration-exploitation balance is difficult to determine from prior knowledge. Additionally, for datasets with rich procedural or descriptive features (e.g., the Membrane dataset), LLM-AL with report-format prompts demonstrates particularly strong performance, suggesting that LLMs can extract latent contextual information and infer relationships that are not easily captured by numerical models alone. Overall, the variability in the number of iterations required by LLM-AL to identify the optimal candidate was comparable to that observed in traditional ML models across different hyperparameter configurations and random seeds, highlighting the potential of LLMs as reliable and robust surrogate models for AL frameworks.

\subsection{Understanding the Acquisition of LLM-AL}
\label{sec:Understanding the Acquisition of LLM-AL}

To better understand how LLM-AL makes acquisitions compared to traditional ML models, we analyzed model trajectories in the standardized feature space. This visualization enables tracking of how different algorithms navigate the search space, whether by following a directed optimization trend or by performing more stochastic movements. As shown in Figure~\ref{fig:AL_distance.png}, we quantified the cumulative distance traveled in the feature space until the optimal candidate was identified.

Across all datasets, the random walk baseline traverses the greatest cumulative distance, exceeding those of every model-based method. GPR consistently travels the shortest distances, likely due to its smooth surrogate and its tendency to converge early toward perceived high-performing regions. GPR and BNN generally exhibit clear convergence behavior, with their cumulative distances flattening as they settle near the optimal candidates. In contrast, LLM-AL generally travels longer distances and shows little evidence of convergence across all datasets: its cumulative trajectory distance increases almost linearly over iterations, suggesting a more discontinuous and exploratory acquisition pattern rather than stabilization near the optima. The tree-based models (i.e., RFR, XGB) often travel distances comparable to those of the LLM, despite relying on fundamentally different mechanisms. This similarity may stem from the highly discontinuous, plateau-like prediction surfaces characteristic of tree-based models, which may produce abrupt jumps between distant regions of the search space. Some representative trajectories are shown in the Supplementary Information in Figure~S2.

Interestingly, despite this apparent “non-convergent” behavior, LLM-AL often reaches optimal targets in fewer iterations than the other models, indicating that the longer distances traveled do not necessarily imply poor performance. In the Membrane dataset, for example, the LLM moves as extensively as a random walk yet is able to identify the optimal targets far sooner than the ML models and random walk. This suggests that its acquisitions may not be driven by learned feature correlations, but perhaps by latent contextual understanding and reasoning patterns inherent to the pretrained model and prior observations. Rather than directly modeling functional mappings, the LLM potentially leverages semantic priors and global contextual cues embedded in its training corpus, allowing it to identify high-performing candidates through a different and potentially more abstract search mechanism. These interpretations remain hypothesis-driven and require further validation to confirm whether these behaviors truly reflect contextual reasoning or are instead the result of stochastic effects. 

\section{Discussion}
We introduced an LLM-based active learning (LLM-AL) workflow that suggests experiments directly from tabular parameters or textual reports and compared it against traditional ML-driven AL models with UCB acquisition across four heterogeneous datasets (metals \& polymers; compositional \& experimental; maximization \& minimization). Overall, LLM-AL reached the optimal candidate in fewer iterations than traditional models in the majority of test scenarios, often requiring less than 30\% of the data. Additionally, unlike conventional surrogate models that require repeated retraining and hyperparameter optimization, LLM-AL is training-free, operating with fixed pretrained weights and relying solely on prompt engineering to guide experimental selection across diverse tasks. 

We designed and evaluated two distinct prompting strategies to understand how LLMs interact with different types of input data: 1) Parameter-format prompts, which provide concise numerical inputs, were highly effective for high-dimensional datasets such as matbench\_steels, where verbose text descriptions would overwhelm the LLM and divert its attention. This compact representation minimizes computational overhead and enables faster, more accurate convergence. 2) Report-format prompts, which excelled in datasets with fewer input variables but rich procedural context, such as the Membrane dataset. In these cases, descriptive text may allow the LLM to uncover implicit relationships and leverage latent domain knowledge that would be lost in purely numerical formats. 

Our analysis of variability shows that LLM-AL remains reliable under both repeated runs and different seeds. Variation is present, especially for longer report-format prompts, but it is generally comparable to the variation observed in ML models across random seeds and hyperparameters. Trajectory analysis revealed that LLM-AL exhibits a highly exploratory search pattern, traversing longer distances in feature space. This behavior may arise from the model’s reliance on semantic cues beyond explicit feature correlations, though the exact mechanism warrants further investigation. 

We find that LLM-AL is particularly effective in problems with low-to-moderate dimensional search spaces where experimental conditions can be meaningfully described in natural language, allowing the model to exploit procedural and contextual information unavailable to numerical surrogates. However, high-dimensional settings with large parameter spaces may still favor compact parameter prompts or conventional ML approaches that can better handle structured feature sets. Given its strong and consistent performance across diverse datasets, LLM-AL may serve not only as a useful baseline for future active learning studies in materials science, but also as a generalizable framework for resource-efficient scientific discovery. 

\section{Methods}

\begin{figure*}[!ht]
\centering
\includegraphics[width=0.75\linewidth]{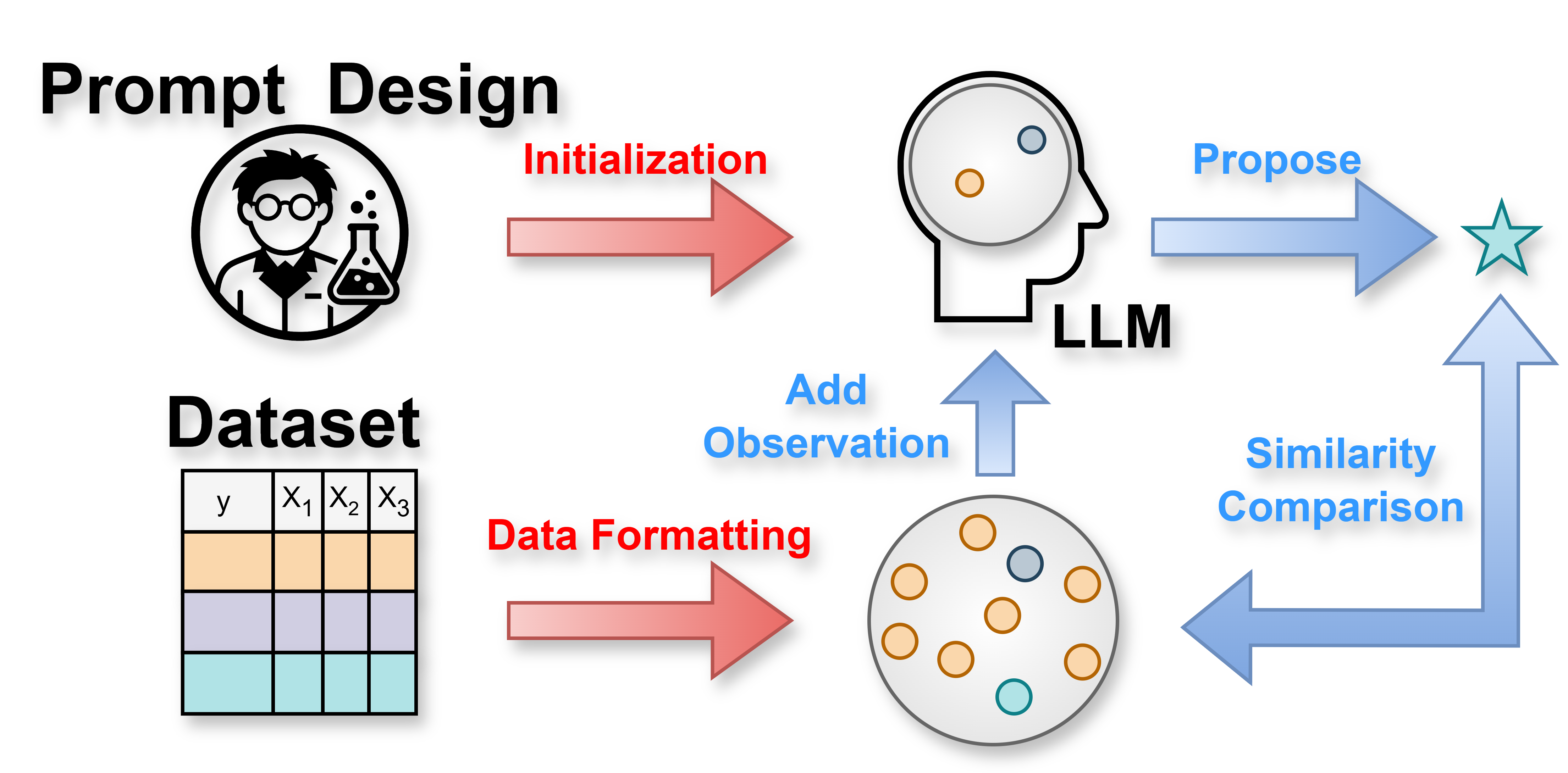}
\caption{Workflow of the LLM-AL framework. The process begins with initialization, where a dataset-specific prompt is designed to provide context and the raw experimental data are formatted into text that the LLM can interpret. At each iteration, the LLM proposes a candidate experiment. The proposed candidate is matched to the closest entry in the data pool using the Cohere Rerank model, and the selected point is added to the labeled dataset. This updated dataset is then fed back into the LLM to guide subsequent selections, creating a closed-loop, iterative process until the best candidate has been identified.}
\label{fig:LLM_workflowDiagram.png}
\end{figure*}

\subsection{Models}
We evaluated the performance of both LLMs and traditional ML models within a pool-based AL framework. The LLM used in this study is \texttt{claude-3.7-sonnet} \cite{Claudesonnet} at $temperature = 0$. We selected \texttt{claude-3.7-sonnet} for its strong reasoning ability and broad scientific knowledge base, which make it well suited for few-shot scientific inference without task-specific fine-tuning \cite{Young_2025, rein2023gpqagraduatelevelgoogleproofqa}. For LLM response matching, we use \texttt{Cohere Rerank-v3.5} \cite{Coherererank} as a similarity-based annotator because rerank models have been shown to outperform embedding-based similarity measures in identifying semantically relevant text pairs \cite{Reimers_Shi_Fayoux_Choi_2023}. The traditional ML baselines include:
\begin{itemize}
    \item Random Forest Regressor (RFR) with 400 trees and all other default hyperparameters in scikit-learn version 1.6.1. 
    \item eXtreme Gradient Boosting (XGBoost) with 400 trees and all other default hyperparameters in version 2.1.4 \cite{XGBOOST}. 
    \item Gaussian Process Regressor (GPR) with kernel as the sum of scaled radial basis function (RBF) kernel and white noise kernel. Hyperparameters found with default L-BFGS-B optimization algorithm in scikit-learn version 1.6.1. The ranges of the hyperparameters to be found when maximizing the log- marginal likelihood are $[1\times10^{-5},1\times10^{5}]$ for the scaling constant in the scaled RBF, $[1\times 10^{-3}, 1\times 10^{3}]$ for the RBF length scale and $[1\times 10^{-3}, 1\times 10^{6}]$ for the noise level.
    \item Bayesian Neural Network (BNN) with fully factorized Gaussian 
posteriors over the weights and biases \cite{blundell2015weightuncertaintyneuralnetworks}, i.e., a mean–field variational approximation. The network consists of 5 hidden Bayesian Linear layers of width 64 neurons each with ReLU activation, and a final Bayesian output layer producing a one-dimensional regression output. The used number of epochs are 1000 with Adam optimizer and learning rate $1\times 10^{-3}$. The Loss function used for training is the sum of Mean Squared Error and Kullback-Leibler divergence. Further information regarding the model implementation is provided in the Supplementary Information in Section~S1.
\end{itemize}

\subsection{Datasets}
We evaluate our AL workflows on four benchmark datasets spanning diverse domains in materials science. These datasets were selected to cover different levels of complexity and represent diverse subfields, providing a robust basis for evaluating the performance of the LLM as a surrogate model compared with traditional ML approaches. The datasets include: 1) \textbf{matbench\_steels} \cite{Dunn2020}: A dataset from Matbench designed to predict the yield strength of steel alloys using elemental composition inputs. 2) \textbf{P3HT/CNT}: A polymer nanocomposite dataset of poly(3-hexylthiophene) and carbon nanotube thin films, designed for optimizing electrical conductivity through variations in compositions \cite{Bash2021, liang_benchmarking_2021}. 3) \textbf{Perovskite}: A dataset targeting the instability index of mixed-cation halide perovskites under heat, humidity, and illumination stress \cite{SUN20211305, liang_benchmarking_2021}. 4) \textbf{Membrane}: A dataset of porous polymeric membranes fabricated via non-solvent induced phase separation under varying polymer concentrations and processing conditions, optimizing for the mechanical properties \cite{wang2025developingvalidatinghighthroughputrobotic}. Additional details of each dataset are summarized in Table~\ref{tab:dataset_summary_all}. It should be noted that many LLMs are pretrained on broad scientific corpora, and their exact training contents are often not (or impossible to be) publicly disclosed. Therefore, it is possible that some of the benchmark datasets used in this study, particularly matbench\_steels, may appear in the scientific literature and thus be included in the model’s pretraining data. The Membrane dataset used in this work, however, is fully original and has never appeared in prior publications, ensuring that no component of the LLM’s training data could include or resemble these experiments.

\begin{table}[!ht]
\centering
\small
\setlength{\tabcolsep}{3pt}
\renewcommand{\arraystretch}{1.15}
\begin{tabularx}{0.48\textwidth}{p{0.17\textwidth} c X c}
\toprule
\textbf{Dataset} & \textbf{Size} & \textbf{Optimization Target (y)} & \textbf{Goal} \\
\midrule
\textbf{matbench\_steels} & 312 & Yield Strength & Max. \\
\textbf{P3HT/CNT} & 323 & Electrical Conductivity & Max. \\
\textbf{Perovskite} & 139 & Instability Index & Min. \\
\textbf{Membrane} & 73 & Elastic Modulus & Max. \\
\bottomrule
\end{tabularx}
\caption{Summary of datasets used for active learning experiments.}
\label{tab:dataset_summary_all}
\end{table}

\subsection{Pool-based Active Learning}

The workflow for the pool-based AL setup is illustrated in Figure~\ref{fig:LLM_workflowDiagram.png}. All AL trials used a batch size of one, meaning that one data point was added to the training data during each iteration. For each model, we tested five different random seeds that determined the initial observations to probe the effect of the initial training data points and model initialization. For traditional ML baselines, we also probed hyperparameter sensitivity by evaluating a fixed set of configurations for each random seed. 

\subsubsection{LLM-Based Active Learning}

For the LLM-based AL approach, each set of candidate input parameters was converted into a single text string. Two input formats were evaluated: (1) a raw parameter string and (2) a more descriptive experimental report generated by the LLM. For each dataset, we designed a prompt that includes the dataset context, the optimization objective, and the observed data points (updated iteratively as few-shot examples). The prompts were designed to provide relevant domain background, narrow the applicable knowledge scope, constrain the feasible input space, specify the optimization target, and instruct the LLM to propose the next experimental condition.

The AL loop begins with a randomly selected seed point, fixed across all models for reproducibility and direct comparison. In each iteration, the LLM proposes a new experimental input. Since this is a pool-based setup, no additional human labeling is required. Instead, the proposed input string is matched to the closest data point in the dataset using \texttt{Cohere Rerank-v3.5}. The matched point is returned with its known target value and then added to the observed dataset. This loop continues until the optimized target value (i.e., the dataset-specific maximum or minimum) is found. The corresponding similarity scores between the LLM-generated suggestions and the matched experiments are reported in the Supplementary Information in Figure~S1. The AL trajectory and the number of iterations taken to reach this stopping criterion are recorded.

\subsubsection{Traditional ML-Based Active Learning}

For the traditional ML-based AL benchmarks, we used the upper confidence bound (UCB) acquisition function across all models. In each iteration, the following steps are performed: 1) standardize inputs using the current labeled pool, 2) train or update the model on the observed data, 3) predict the mean and uncertainty over the unlabeled pool, 4) compute the UCB value for each candidate as $\mu + \alpha\sigma$, where $\mu$ is the predicted mean, $\sigma$ is the model uncertainty, and $\alpha$ controls the exploration–exploitation balance, and 5) select the point with the highest UCB score.

The AL loop terminates once the optimized target value is reached. We log the full AL trajectory for each model and compare the number of iterations required to achieve optimal performance across datasets and methods. For existing dataset benchmarks (i.e., P3HT/CNT and Perovskite datasets), ML-based AL models in the literature have been reported to identify the optimal point using about 40–-60\% of the available data \cite{liang_benchmarking_2021}. The ML baselines in this study were confirmed to reach comparable benchmark performance to ensure a fair comparison with the LLM-based approach.

\section*{Author Contributions}
H.W. and E.K. conceived and designed the project. E.K., and J.H.S. supervised the project. H.W. and R.E.C. conducted the experiments and drafted the manuscript. J.W. provided technical support for the collection of the Membrane dataset. H.W., R.E.C., J.W., Y.F., E.K., and J.H.S. discussed the results. All authors reviewed and edited the manuscript.

\section*{Competing Interests}
The authors declare no conflicts of interest.

\section*{Data and Code Availability}
Data and code will be made available on GitHub. 

\section*{Acknowledgment}
The authors acknowledge financial support from the Natural Sciences and Engineering Research Council of Canada (NSERC) Alliance grants (ALLRP 601812-24), the National Research Council of Canada’s Critical Battery Materials Initiative (CBMI-002-1). The research was also, in part, made possible thanks to funding provided to the University of Toronto’s Acceleration Consortium by the Canada First Research Excellence Fund (CFREF-2022-00042).

\begingroup
\footnotesize
\bibliography{main} 
\bibliographystyle{rsc} 
\endgroup

\end{document}


\begin{center}
{\Large \textbf{Supplementary Information}}\\
\end{center}

\subsection{Bayesian Neural Network Loss Derivation and Training Procedure}\label{fig:SI_bayesian}
We consider a regression model of the form:
\[
y = f(x; W, b) + \epsilon, \quad \epsilon \sim \mathcal{N}(0, \sigma_y^2)
\]
where \( f(x; W, b) \) is a neural network with weights \( W \) and biases \( b \), and \( \epsilon \) is Gaussian observation noise with variance \( \sigma_y^2 \). Under this model, the log-likelihood of a single observation becomes:
\[
\begin{aligned}
\log p(y \mid x, W, b) = 
& -\frac{1}{2\sigma_y^2} \| y - f(x; W, b) \|^2 \\
& - \frac{1}{2} \log(2\pi \sigma_y^2)
\end{aligned}
\]
Since the second term is constant, and we fix \( \sigma_y = 1 \), maximizing the log-likelihood is equivalent to minimizing the mean squared error (MSE) scaled by $\frac{1}{2}$.

\medskip

In the Bayesian framework, rather than learning point estimates for \( W \) and \( b \), we infer a posterior distribution over them. This is approximated using a fully factorized Gaussian variational distribution \( q(W, b) \), and training is done by maximizing the Evidence Lower Bound (ELBO):
\[
\begin{aligned}
\log p(y) \geq\ 
& \mathbb{E}_{q(W,b)}\left[\log p(y \mid x, W, b)\right] \\
& - \mathrm{KL}\left(q(W, b) \,\|\, p(W, b)\right)
\end{aligned}
\]
Therefore, the training loss is defined as the negative ELBO, which results in the following objective:
\[
\mathcal{L} =\frac{1}{2}\mathrm{MSE}(y, \hat{y}) + \mathrm{KL}\left(q(W, b) \,\|\, p(W, b)\right)
\]
where:
\begin{enumerate}
\item \( \hat{y} = f(x; W, b) \) is the prediction,
\item KL is the Kullback–Leibler divergence between the approximate posterior \( q(W, b) = \mathcal{N}(\mu, \sigma^2) \) and the standard Gaussian prior \( p(W, b) = \mathcal{N}(0, 1) \).
\end{enumerate}

The KL divergence has the closed-form expression:
\[
\mathrm{KL}(q \,\|\, p) = \log\left(\frac{1}{\sigma}\right) + \frac{\sigma^2 + \mu^2 - 1}{2}
\]

\medskip

Each linear layer in the neural network is replaced by a Bayesian linear layer, where weights and biases are sampled as:
\[
W \sim \mathcal{N}(\mu_W, \sigma_W^2), \quad b \sim \mathcal{N}(\mu_b, \sigma_b^2)
\]
The parameters \( \mu_W, \mu_b \) are initialized uniformly in \( [-0.2, 0.2] \), and the standard deviations are initialized with \( \sigma_a=\sigma_b = \exp(-5) \approx 6.74 \times 10^{-3} \).
Training is performed using the Adam optimizer. To estimate predictive uncertainty at inference time, we use Monte Carlo sampling with \( 1000 \) forward passes and compute the empirical mean and standard deviation of the predictions.

\clearpage
\subsection{Rerank Similarity Scores of the Proposals to the Pool Data}
\begin{figure*}[!ht]
\centering
\includegraphics[width=\linewidth]{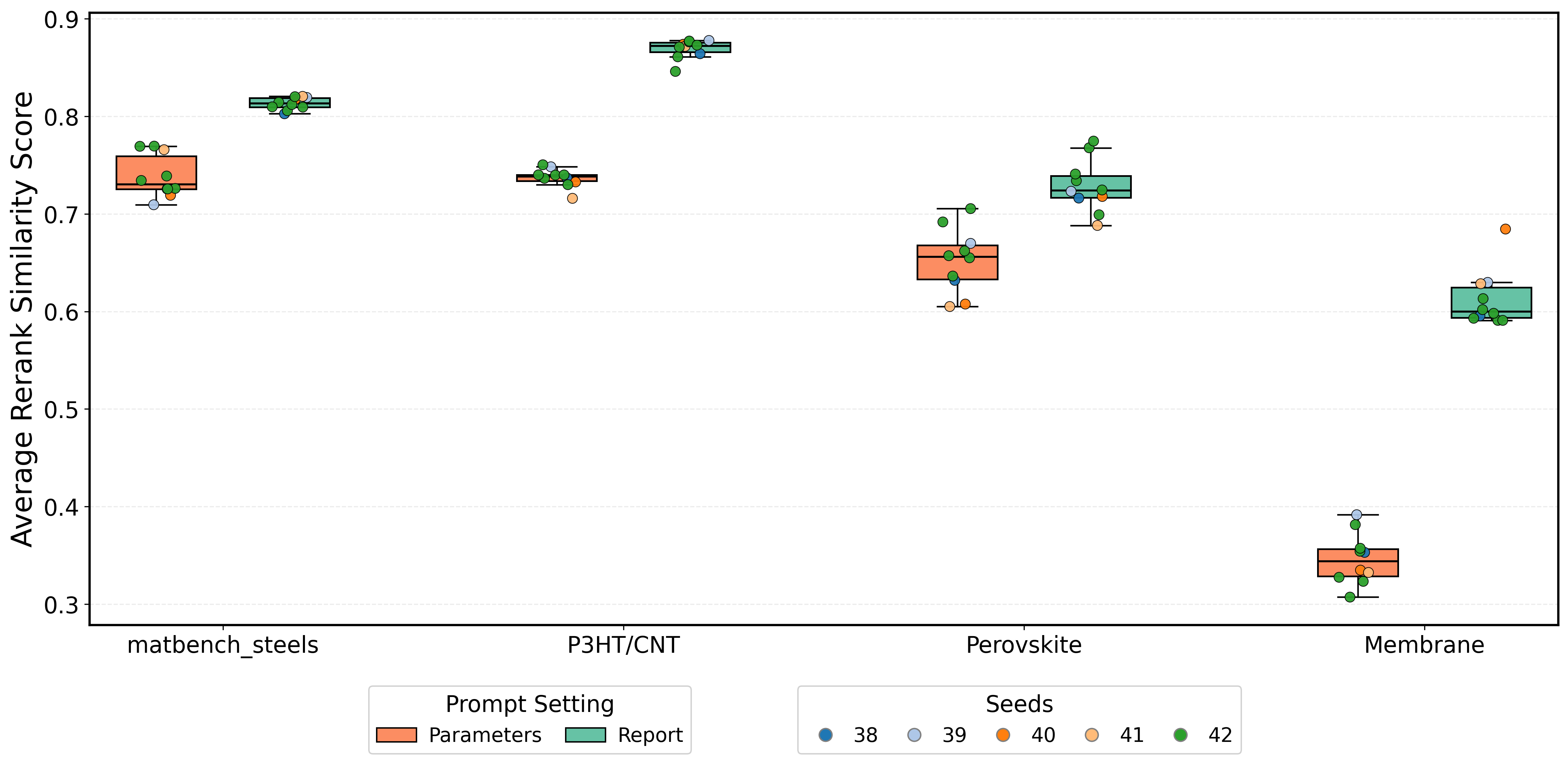}
\caption{Average rerank similarity scores across datasets and prompt settings.
Each box represents the distribution of mean Cohere ReRank v3.5 similarity scores computed per trajectory. The orange and green boxes correspond to the “Parameters” and “Report” prompting modes, respectively, while colored markers denote results from different random seeds.
Across all datasets, Report-based prompts generally produce higher semantic consistency with prior experimental context. Noticeably lower similarities are observed for the parameter-format condition in the Membrane dataset.}
\label{fig:AL_llm_similarity_SI}
\end{figure*}

\clearpage
\subsection{Active Learning Trajectories in PCA Space}
\begin{figure*}[!ht]
\centering
\includegraphics[width=0.95\linewidth]{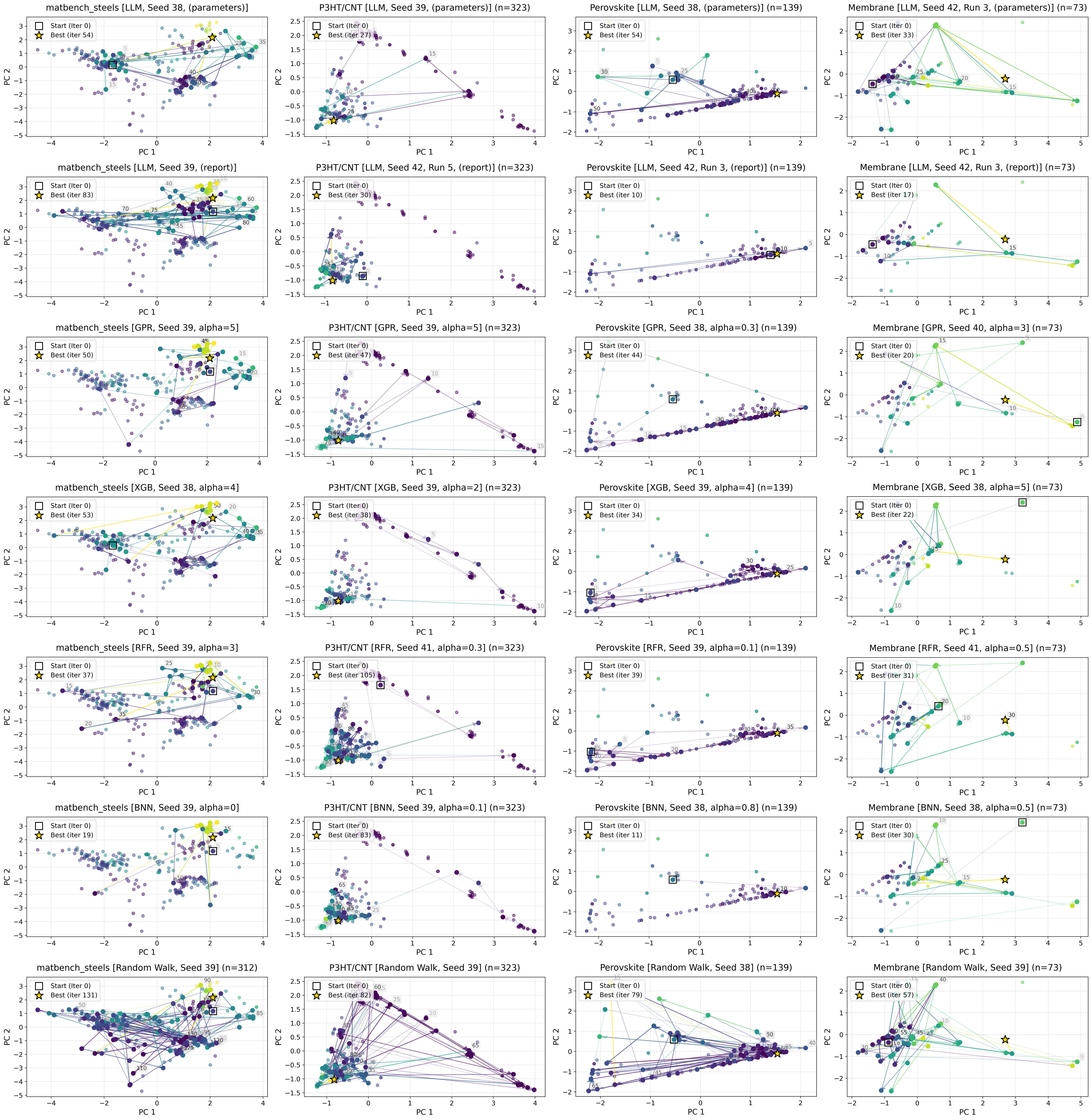}
\caption{Principal Component Analysis (PCA)‐projected trajectories of all active learning and baseline runs across datasets. Each point represents an experimental candidate projected into the first two principal components of the standardized feature space, with color indicating the observed target value. Lines trace the sequential exploration path of each model, beginning with the square marker (Iteration 0) and culminating at the best-performing sample (star marker). Separate panels correspond to LLM-AL using either parameter- or report-format prompts, traditional ML baselines (i.e., GPR, RFR, XGB, and BNN), and random walk baselines. The trajectories illustrate how models traverse the feature space in an active learning run.}
\label{fig:AL_pca_SI}
\end{figure*}